%% file: main.tex
\documentclass[10pt,twocolumn,letterpaper]{article}

\usepackage{cvpr}              

\input{preamble}

\definecolor{cvprblue}{rgb}{0.21,0.49,0.74}
\usepackage[pagebackref,breaklinks,colorlinks,citecolor=cvprblue]{hyperref}

\def\paperID{****}
\def\confName{CVPR}
\def\confYear{2024}

\title{R2-Talker: Realistic Real-Time Talking Head Synthesis with Hash Grid Landmarks Encoding and Progressive Multilayer Conditioning}
\author{Zhiling Ye \quad LiangGuo Zhang \quad Dingheng Zeng \quad Quan Lu \quad Ning Jiang\\
Mashang Consumer Finance Co., Ltd. \\
}

\begin{document}
\maketitle
\input{sec/0_abstract}    
\input{sec/1_intro}

\input{sec/2_experiments}
\input{sec/3_finalcopy}
{
    \small
    \bibliographystyle{ieeenat_fullname}
    \bibliography{main}
}
\input{sec/X_suppl}

\end{document}

%% file: preamble.tex
\usepackage[dvipsnames]{xcolor}


%% file: sec/0_abstract.tex
\begin{abstract}
\vspace{-0.3cm}
Dynamic NeRFs have recently garnered growing attention for 3D talking portrait synthesis. Despite advances in rendering speed and visual quality, challenges persist in enhancing efficiency and effectiveness. We present R2-Talker, an efficient and effective framework enabling realistic real-time talking head synthesis. Specifically, using multi-resolution hash grids, we introduce a novel approach for encoding facial landmarks as conditional features. This approach losslessly encodes landmark structures as conditional features, decoupling input diversity, and conditional spaces by mapping arbitrary landmarks to a unified feature space. We further propose a scheme of progressive multilayer conditioning in the NeRF rendering pipeline for effective conditional feature fusion. Our new approach has the following advantages as demonstrated by extensive experiments compared with the state-of-the-art works: 1) The lossless input encoding enables acquiring more precise features, yielding superior visual quality. The decoupling of inputs and conditional spaces improves generalizability. 2) The fusing of conditional features and MLP outputs at each MLP layer enhances conditional impact, resulting in more accurate lip synthesis and better visual quality.  3) It compactly structures the fusion of conditional features, significantly enhancing computational efficiency.

\end{abstract}

%% file: sec/1_intro.tex


\section{Introduction}
\label{sec:intro}

Audio-driven talking portrait synthesis has seen growing attention from both academia and industry, with not only various applications like in virtual characters, virtual assistants, and video games but also significant research values in the generative AI era \cite{zhan2023multimodal}. Over the past few years, many researchers have explored various solutions, typically based on deep generative models \cite{method2:prajwal2020lip, method1:thies2020nvp, lu2021live}, to this task. For instance, Generative Adversarial Networks (GAN) \cite{goodfellow2020generative} have been employed from only predicting the audio-synchronized lip movements \cite{method2:prajwal2020lip} to generating the whole face \cite{sun2021speech2talking}. However, the GAN-based approach suffers from some limitations such as inherently unstable training, and mode collapse.

Recently, Neural Radiance Fields (NeRF) \cite{mildenhall2020nerf} have gained popularity in talking portrait synthesis. Various approaches have been explored, taking 3D Morphable Models \cite{method4:gafni2021dynamic}, audio \cite{guo2021adnerf, tang2022radnerf, li2023ernerf}, and landmarks \cite{ye2023geneface, ye2023geneface++} as condition. However, these methods do not have enough investigations of the input conditions of NeRF-based audio portrait synthesis. Notably, AD-NeRF \cite{guo2021adnerf} employs conventional audio processing techniques to derive conditional features, while RAD-NeRF \cite{tang2022radnerf} and ER-NeRF \cite{li2023ernerf} leverage more streamlined architectural innovations to enhance condition effects. Similarly, Geneface \cite{ye2023geneface} and other approaches utilize landmarks as conditions but persist with processing methodologies akin to RAD-NeRF. These pre-processing approaches for conditions necessitate a multitude of filtering operations, such as the sliding window mechanism common in audio processing. Moreover, the compression of high-dimensional conditions into exceedingly low dimensions, as witnessed in HyperNeRF \cite{park2021hypernerf}, regrettably leads to the loss of substantial informational content. Consequently, it results in a suboptimal outcome during the NeRF phase, impeding the realization of optimal rendering results.

\begin{figure}
    \centering
    \includegraphics[width=1\linewidth]{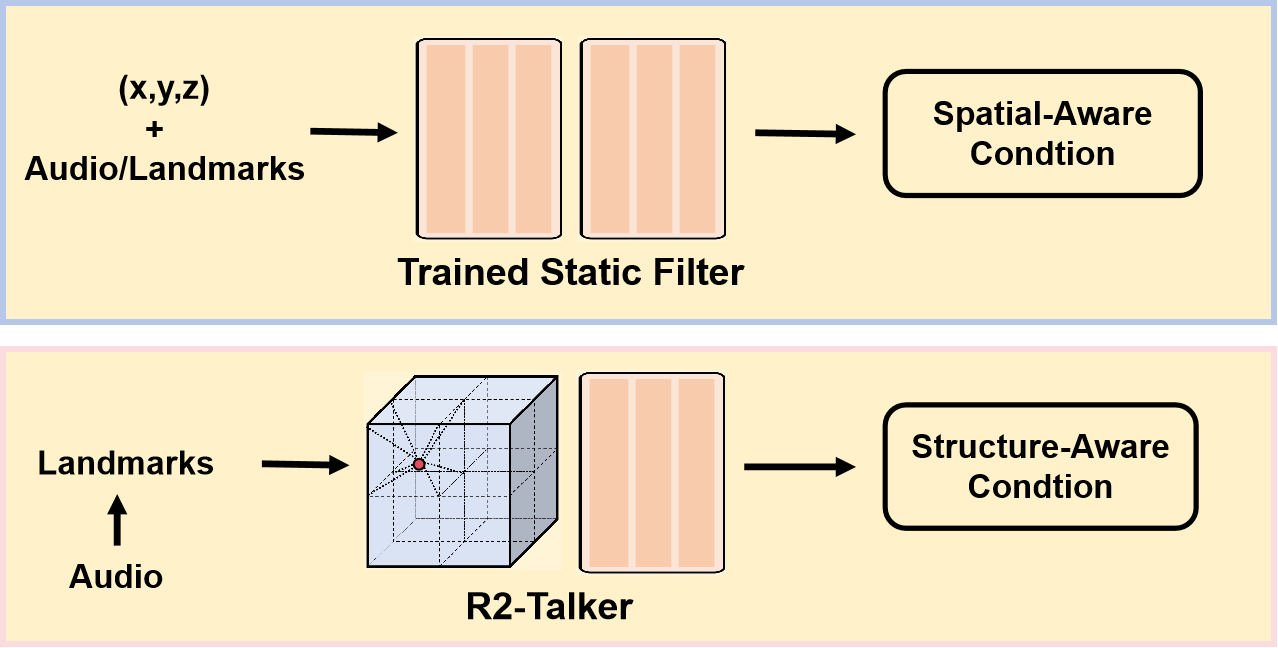}
    \caption{Instead of extracting position and audio/landmark information through a parameterized static filtering (feature extractor) for spatial-aware conditions as used in previous methods, our R2-Talker implements a dynamic filtering system, taking landmarks as input and passing them through a hash grid encoder and an MLP, to extract structure-aware conditions. }
    \label{fig:R2-Talker Express}
\end{figure}

In this paper, we propose a novel framework, named \textit{R2-Talker, to further investigate the impact of the condition.} For a quick understanding of the core idea, please refer to Fig. \ref{fig:R2-Talker Express}. This framework, designed with a compact pipeline, significantly enhances rendering speed and improves the fidelity of the generated results. Specifically, R2-Talker adopts the motion generator of GeneFace++ \cite{ye2023geneface++} which generates 3D landmarks from input driving audio. Then, instead of processing with a feature extractor, R2-Talker encodes the raw 3D facial landmarks with a structure-aware encoder to generate more precise and generalized conditional features. Here, the encoder is based on hash grids which is proposed in instant-NGP \cite{mueller2022instant}. In RAD-NeRF \cite{tang2022radnerf}, the audio-spatial condition calculation, performed on a per-sample ray basis, is resource intensive; while R2-Talker directly applies the conditional feature as a global condition for all position features in NeRF. In addition, inspired by StyleGAN \cite{karras2019style}, R2-Talker introduces a multi-layer approach to improve the fusing efficiency of conditional features and positional features. Furthermore, inspired by Neuralangelo \cite{li2023neuralangelo}, R2-Talker proposes the coarse-to-fine training technique to further enhance the rendering quality. 

In summary, the main contributions of our work are as follows: 
\begin{itemize}
    \vspace{0.1cm}
    \item We propose a structure-aware method for encoding audio-driven 3D facial landmarks based on multi-resolution hash grids. Compared to previous approaches, this achieves lossless encoding and a decoupled conditional feature space, enabling better quality and generalization.
    \vspace{0.1cm}
    \item We propose a new conditioning method for fusing conditional features and positional features. Building on this method, we further introduce progressive optimization into the training process.
    \vspace{0.1cm}
    \item Extensive experiments show that our method can achieve significantly better performance in terms of rendering quality, rendering speed, and audio-lip synchronization compared with the state-of-the-art works.
\end{itemize}

\section{Related Work}
\begin{figure*}
    \centering
    \vspace{-0.2cm}
    \includegraphics[width=1.0\linewidth]{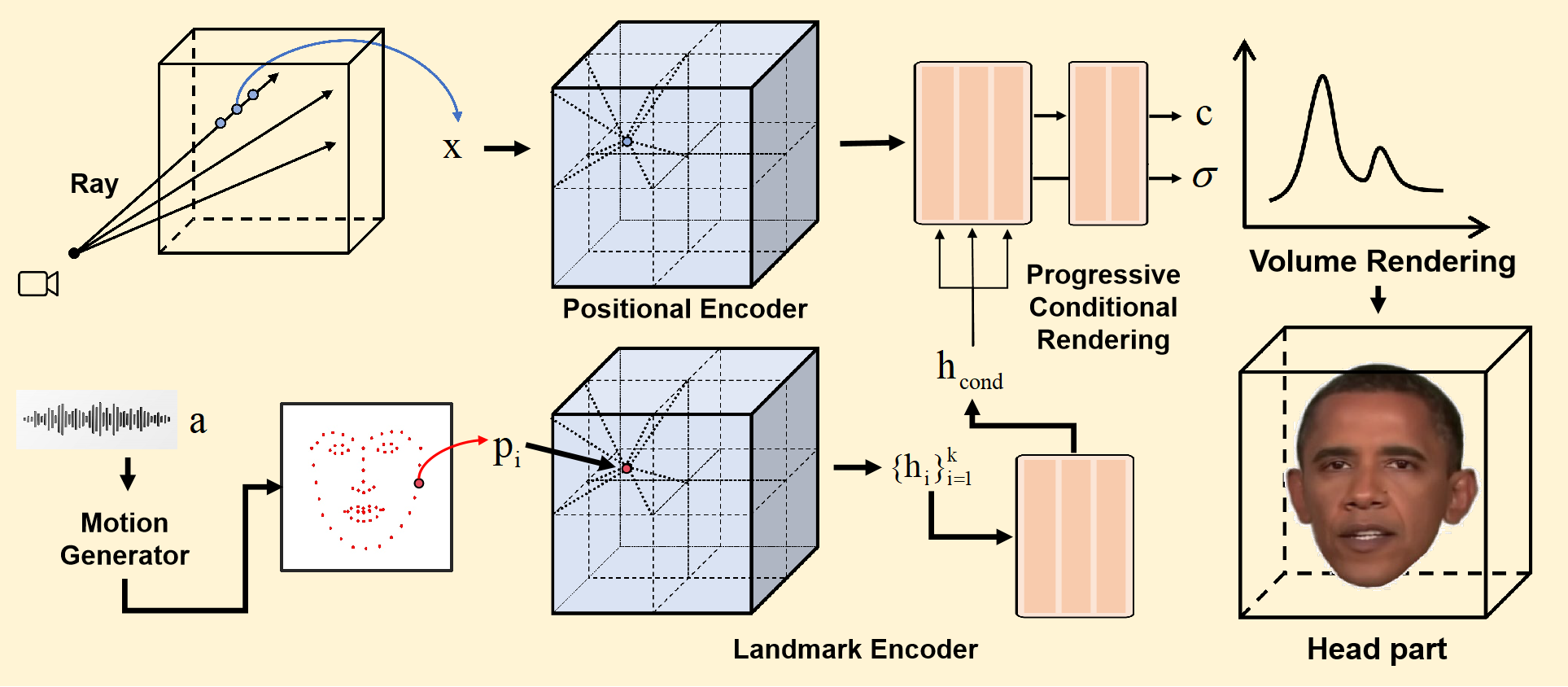}
    \caption{\textbf{Overview of R2-Talker.} Firstly, the driving audio is input into the \textbf{Motion Generator} to generate corresponding 3D facial landmarks, which are then used to generate conditional features by the \textbf{Landmark Encoder}. These features are input into the \textbf{Progressive Conditional Rendering} module to obtain emitted color and density. Finally, the head part is generated through \textbf{Volume Rendering}. Parallel positions sampled along the ray are input into the \textbf{Positional Encoder} and then fused with conditional features. For the torso part, we adopt the torso-generating module from RAD-NeRF \cite{tang2022radnerf} and omit details here for simplicity.}
    \label{fig: pipeline}
    \vspace{-0.5cm}
\end{figure*}
\subsection{Audio-driven Talking Portrait Synthesis} 
Audio-driven talking portrait synthesis is to visually reinterpret a specific individual from any input driving audio. Various methods have been proposed aiming to generate realistic and accurate audio-lip synchronized talking portrait videos. We categorize these methods into traditional methods, deep learning methods, and model-based methods. Traditional methods \cite{method1:bregler2023video,method1:thies2020nvp} primarily establish phoneme-mouth shape correspondence rules and employ stitching-based techniques to modify the mouth shape. Deep learning based methods  \cite{method2:chung2017you,method2:ji2021audio,method2:cudeiro2019capture,method2:pham2017speech,method2:prajwal2020lip,method2:song2018talking,method2:taylor2017deep,method2:vougioukas2020realistic,method2:wang2021audio2head,method2:zhang2021flow,method2:zhou2019talking,method2:zhou2021pose,method2:guan2023stylesync} directly generates images corresponding to the input audio. However, these methods heavily rely on the resolution of the input data and are processed in 2D space, resulting in their limitations, e.g., the generation of images with a fixed resolution, and the inability to control head posture. Model-based methods \cite{method3:zhou2020makelttalk,method3:yi2020audio,method3:wu2021imitating,method3:wang2020mead,method3:suwajanakorn2017synthesizing,method3:song2022everybody,method3:meshry2021learned,method3:gao2022reconstructing,method3:das2020speech,method3:chen2019hierarchical,method3:chen2020talking} primarily enhance talking synthesis portraits by utilizing structural representations such as facial landmarks and 3D malleable facial models. However, these intermediate representation processes inevitably introduce new errors, which can affect the final output quality.

\subsection{NeRF-based Talking Head Synthesis}
NeRF is a method to represent scenes as neural radiation fields, enabling view synthesis and 3D reconstruction. It excels at generating realistic 3D scene models from 2D images and has a broad range of potential applications. Recent studies \cite{method4:gafni2021dynamic,guo2021adnerf,method4:liu2022semantic,method4:shen2022learning,method4:chatziagapi2023lipnerf, method4:yao2022dfa} have leveraged NeRF \cite{mildenhall2020nerf} to talking portraits synthesis. This NeRF-based approach combines deep learning and model-based advantages, offering the advantages of realistic rendering at any resolution, active head posture control, and eliminating intermediate processes. Earlier methods implicitly constructed the head model, resulting in a trade-off between speed, effectiveness, and practicality. With the evolution of NeRF technology, RAD-NeRF \cite{tang2022radnerf} builds upon instant-NGP \cite{mueller2022instant} and has made significant advancements in both efficiency and quality of generated results. Subsequently, ER-NeRF \cite{li2023ernerf} further enhances the efficiency of neural radiation field representation. On the other hand, the GeneFace \cite{ye2023geneface, ye2023geneface++} series of work converts audios into 3D landmarks, which are used as input to NeRF, thereby enhancing the generalizability of the system. However, the mentioned methods do not have enough investigations of the efficient and effective encoding and conditioning of driving inputs. By designing a structure-aware encoder and proposing a progressive conditioning scheme, we introduce a realistic and real-time talking head synthesis method. 

\section{Method}
R2-Talker adopts the audio-to-3D landmark conversion similarly as \cite{ye2023geneface, ye2023geneface++}. Our key contributions lie in the landmark-to-NeRF phase (as shown in Fig. \ref{fig: pipeline}): Landmark Encoding with Hash-Grid (Section \ref{subsec:Landmark Encoding}) and Progressive Multi-layer Conditioning (Section \ref{subsec:Progressive Multi-Layer Conditioning}).

\subsection{Preliminaries and Problem Setting}
\label{subsec:Preliminaries and Problem Setting}
The basic idea of NeRF \cite{mildenhall2020nerf} is to model a scene as a continuous 3D function, which can be formally defined as follows: 
\begin{equation}  
\mathcal{F}: (\textbf{x},\textbf{d}) \rightarrow (\textbf{c}, \sigma),
\end{equation}
where \(\mathcal{F} \) is the scene function and  \( \textbf{x} = (x,y,z)  \) are the 3D coordinates and  \( \textbf{d} \) is the viewing direction. \( \textbf{c} \) represents the emitted color and \( \sigma \) represents the density of the scene respectively at the given point \( \textbf{x} \). \( \textbf{r}(t) = \textbf{o} + t\textbf{d} \) is the ray from camera center \( \textbf{o} \) cross the pixel. Given sequentially sampled points \( \{t_{i}\} \) along the ray \( \textbf{r}(t) \), densities \( \{\sigma_{i}\} \) and emitted colors \( \{\textbf{c}_{i}\} \) can be obtained by \(\mathcal{F} \). The pixel value \( \textbf{C}(\textbf{r}) \) is numerically calculated by:
\begin{equation} 
\textbf{C}(\textbf{r}) = \sum_{i} T_{i} (1-\textbf{exp}(-\sigma_{i}\Delta_{i})) \textbf{c}_{i},
\end{equation}
where \( \Delta_{i} = t_{i+1} - t_{i} \) is the step size and \( T_{i} \), accumulated transmittance from \( t_{0} \) to \( t_{i} \), is computed by:
\begin{equation} 
 T_{i}= \prod_{j<i}\textbf{exp}(-\sigma_{j}\Delta_{j}).
\end{equation}

Based on differentiable voxel rendering processes, the NeRF-based talking portrait generation uses additional input driving audio \( \textbf{a} \), along with \( \textbf{x} \) and \( \textbf{d} \):
\begin{equation}  
\mathcal{F}: (\textbf{x},\textbf{d}, \textbf{a} )  \rightarrow (\textbf{c}, \sigma).
\end{equation}

In this work, we draw upon concepts from prior NeRF-based studies \cite{tang2022radnerf, ye2023geneface, li2023ernerf, ye2023geneface++} and introduce innovative techniques to enhance efficiency, visual quality and lip-sync accuracy for audio-driven talking portrait synthesis.
 
\subsection{Landmark Encoding with Hash-Grid} 
\label{subsec:Landmark Encoding}
In the audio-driven talking portrait synthesis task, extracting informative and distinguishable conditional features from input driving audio is paramount. In RAD-NeRF \cite{tang2022radnerf}, ER-NeRF \cite{li2023ernerf}, and others, input driving audio passes through a generalized filter \(\mathcal{G} \) (slide windows + ASR + smoothed with 1D convolution + self-attention) to generate conditional features. After training, the filter  \(\mathcal{G} \) is frozen, and the entire feature extraction system is a static nonlinear filtering system. Geneface++ \cite{ye2023geneface++} converts input driving audio to 3D facial landmarks with a motion generator, reshapes the 3D landmarks into a 1D vector, and uses the above-mentioned filter  \(\mathcal{G} \) to extract conditional features, which is also a static nonlinear filtering system.

Once the model is trained, its parameters are frozen and the conditional feature generation pipeline can be viewed as a static nonlinear filter. Although this pipeline demonstrates strong performance on the training and validation sets, practical applications present numerous unsolved challenges. The model often fails to generalize as expected. For instance, compared to the training and validation sets, real-world input audio can span diverse emotions, languages, and genders. By decoupling input driving audio from person-specific conditional features, we propose a structure-aware nonlinear filtering method. Concerning the person-specific nature of our task, we assume all possible conditional features for a person comprise a feature space \(\textbf{H}=\{{\theta}_{i}\}\), where \( \theta_{i}\) is a trainable parameter representing a set of basis vectors. For a given point in feature space, it can be represented by basis vectors and corresponding weights. We represent such a feature space with a multi-resolution hash grid as proposed in Instant-NGP \cite{mueller2022instant}. The entries in the hash grid correspond to basis vectors \( \theta_{i}\), and the weights are calculated based on the distance between the query position of the input and the entry position. This constitutes a queryable continuous feature space:
\begin{figure}
    \centering
    \includegraphics[width=1.0\linewidth]{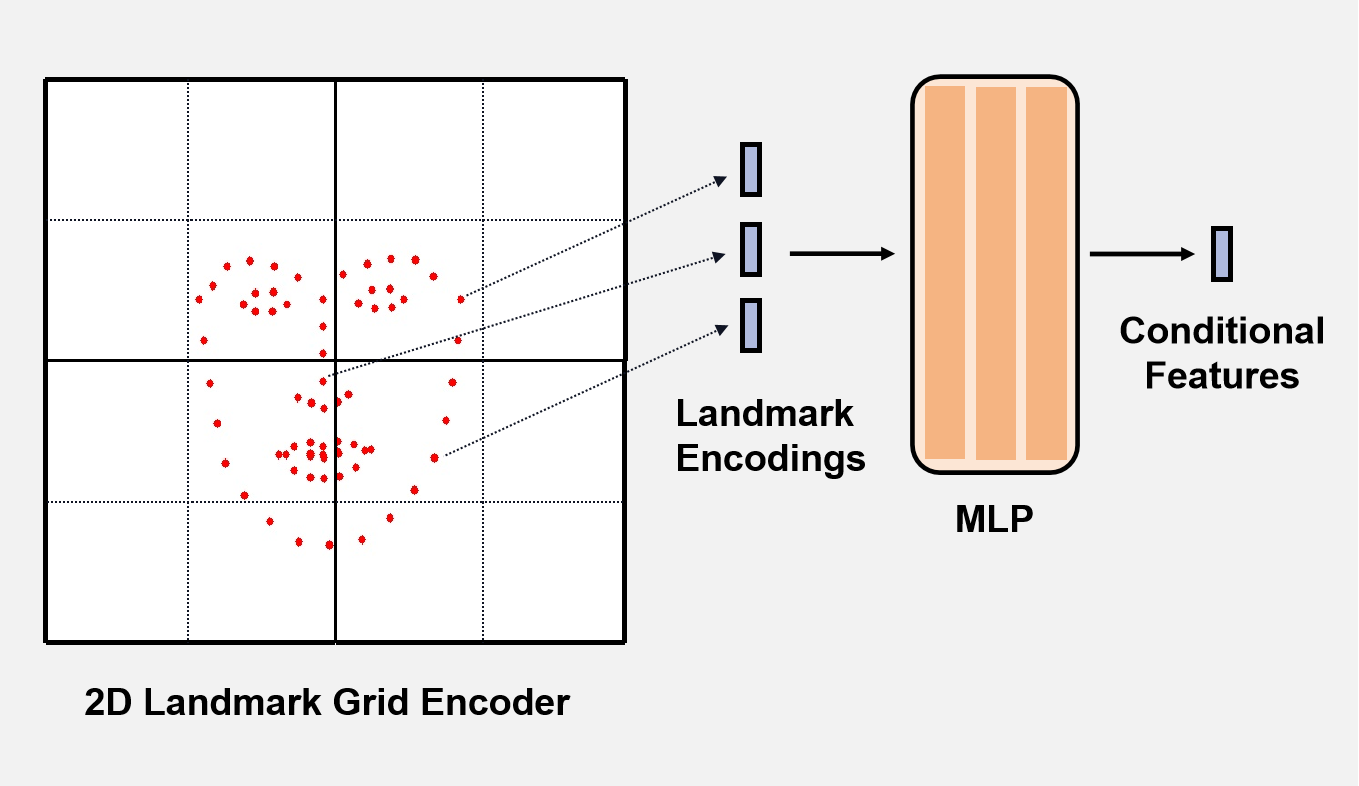}
    \vspace{-0.4cm}
    \caption{\textbf{Landmark Encoder.} The figure depicts a 2D example illustrating the landmark encoder process. Based on landmark (red dots in the grid) coordinates \( (x, y) \), corresponding landmark encodings are extracted from a grid encoder using multi-resolution hash grids. These encodings are concatenated and then passed through a small MLP, yielding conditional features.}
    \label{fig: Landmark encoder}
\end{figure}

\begin{equation}  
\mathcal{H}_{\theta}: \textbf{p}  \rightarrow \textbf{h},
\end{equation}
where \( \textbf{p} \) represents the continuous query coordinate, and \( \mathbf{h} \in \mathbb{R}^{d} \) denotes the corresponding encoding. \( \theta \) signifies trainable feature bases, and \( d \) corresponds to the dimension of the encoding. We propose a dynamic structured filter on \(\mathcal{\textbf{H}}\): 
\begin{equation}  \mathcal{G}\\(\textbf{a}, \mathcal{H}_{\theta} \\):  \textbf{P} \rightarrow  \hat{\textbf{H}}, \end{equation}
where \(\textbf{P}=\{\textbf{p}_j\}(j=1,2,..,k)\), \( \textbf{p}_j \in \mathbb{R}^{3} \) is a coordinate of 3D facial landmark and \( \textbf{a} \) is the input driving audio which is turned to be the 3D facial landmarks by a motion generator. And \( \hat{\textbf{h}} \in \mathbb{R}^{d \times k}  \) is the concatenated landmark encodings, specifically:
\begin{equation} \hat{\textbf{h}} = (  \mathcal{H}(\textbf{p}_1) ,   \mathcal{H}(\textbf{p}_2) , ... ,   \mathcal{H}(\textbf{p}_k) ). \end{equation}
The \( \hat{\textbf{h}} \) would be subsequentially processed by an \textit{MLP}:
\begin{equation}\textbf{h}_{cond} = MLP( \hat{\textbf{h}} ).\label{eq: conditional feature}\end{equation} 
\( \textbf{h}_{cond} \) serves as a conditional feature, which is utilized for controlling subsequent NeRF rendering modules. \( \mathcal{G} (\textbf{a}, \mathcal{H}_{\theta} ) \) losslessly encodes the structure of input driving landmarks into its calculation process. In this process, divergent input driving landmarks are decoupled from the conditional feature, thereby leading to a novel structure-aware nonlinear filtering system. Compared to the previously established methods, our innovative method is capable of encoding input driving information in a lossless manner. Consequently, it generates conditional features with superior representation capabilities, thereby enhancing the rendering effects which can be verified by self-driven experiments (as detailed in Table \ref{tab:self-driven}). Furthermore, with the aid of conditional feature visualization (as shown in Fig. \ref{fig: Conditional Feature Visualization}), it becomes visually evident that our method yields more superior representation and more separable conditional features in comparison to other methods like RAD-NeRF \cite{tang2022radnerf}, Geneface++ \cite{ye2023geneface++}. Additionally, for driving audios that fall outside the scope of our training and validation sets, our method can still extract conditional features from a unified feature space. This key attribute greatly benefits the strong impressive generalization capabilities of our method. This advantage is further demonstrated in our cross-driven experiments, as detailed in Table \ref{tab:cross-driven}.

\begin{figure}
    \centering
    \vspace{-0.4cm}
    \includegraphics[width=1\linewidth]{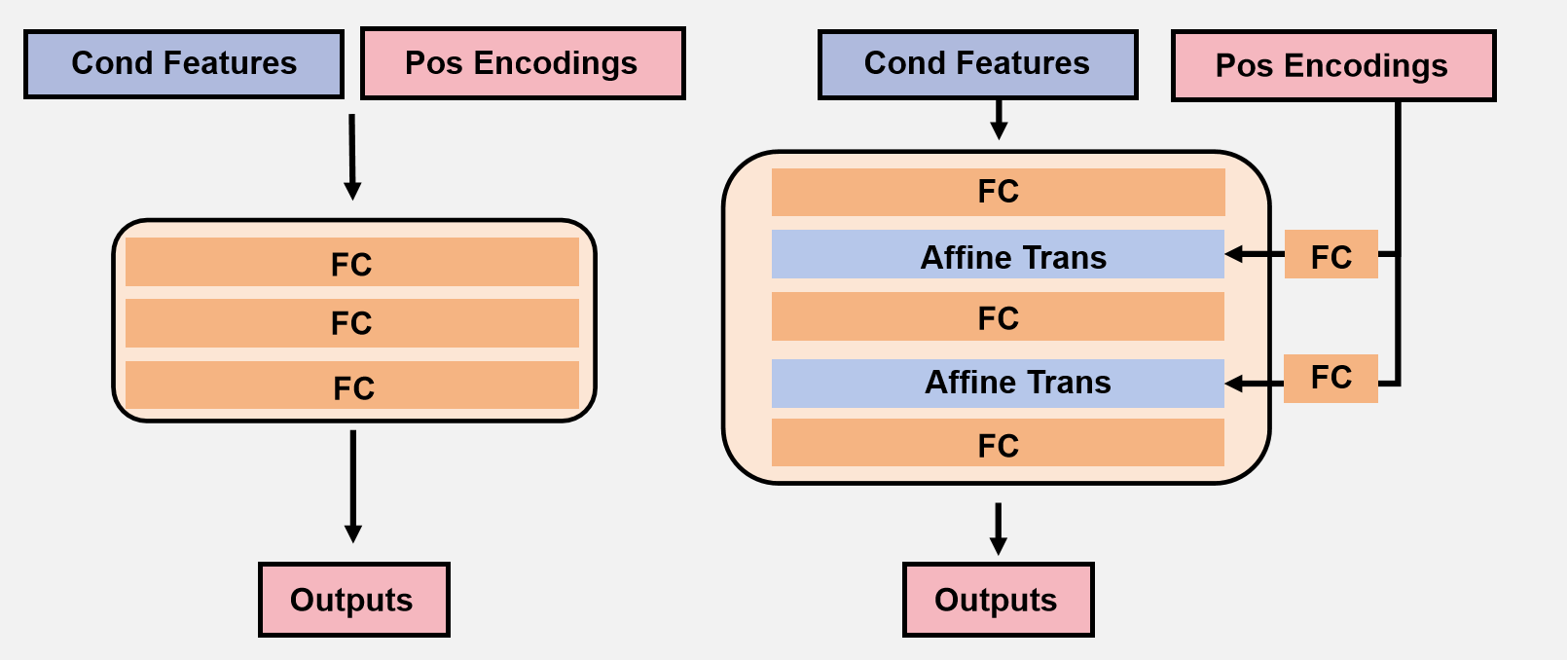}
    \vspace{-0.4cm}
    \caption{\textbf{GLO} (left) and \textbf{M-AGLO} (right). 1) Global latent optimization (GLO) obtains outputs by concatenating conditional and positional features and passing them through an MLP. 2) M-AGLO first transforms the conditional feature through a corresponding fully connected layer, then integrates it with the positional feature through an affine transformation on the intermediate results of each MLP layer. For simplicity, we refer to the conditional and positional features as cond feature and pos feature, and the affine transformation as affine trans. If not specified, these terms will be abbreviated as such in subsequent figures.}
    \label{fig:GLO_and_AGLO}
\end{figure}
Figure \ref{fig: Landmark encoder} shows a 2D landmark encoder example. The red dot on the left represents 2D facial landmarks, and the white grid is the hash grid encoder. Based on coordinates \( (x, y) \), each landmark queries its corresponding encodings from the hash grid encoder. The encodings of all landmarks are concatenated as landmark encodings. These undergo processing by MLP to obtain conditional features. 

\subsection{Progressive Multilayer Conditioning}
\label{subsec:Progressive Multi-Layer Conditioning}
Our proposed progressive multilayer conditioning involves two parts: \textbf{Multilayer Affine Generative Latent Optimization (M-AGLO)} and \textbf{Conditional Progressive Optimization}, which incrementally improves the performance of conditioning.

\begin{figure}
    \centering
    \vspace{-0.4cm}
    \includegraphics[width=1\linewidth]{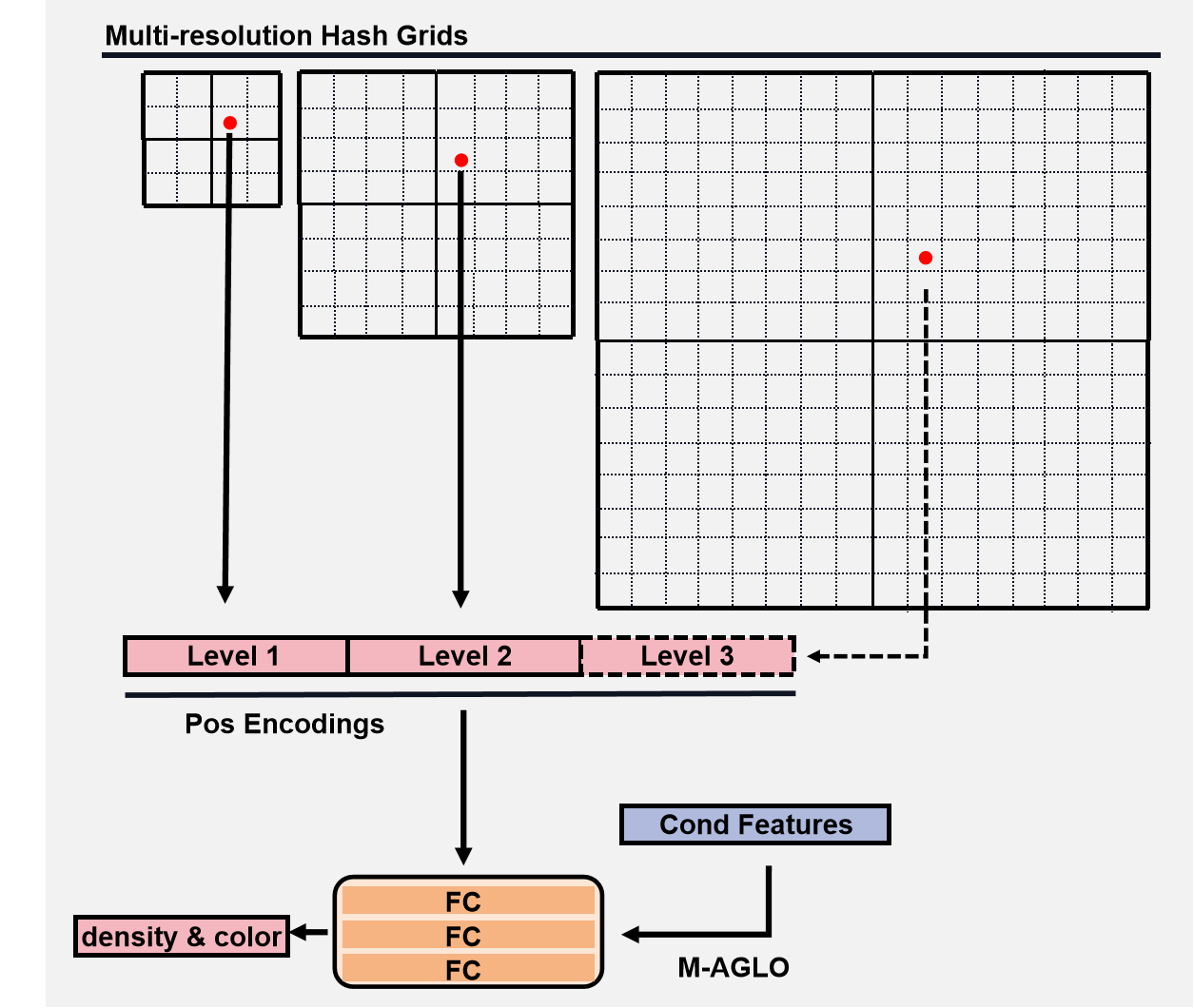}
    \caption{\textbf{Conditional Progressive Optimization.} Positional features are gradually concatenated together as the training steps increase. Conditional features are fused with positional features at the intermediate layer outputs of MLP via M-AGLO. Hence, M-AGLO can adapt to dynamic changes of positional features.} 
    \label{fig: Conditional Progressive Optimizationl}
\end{figure}

\noindent \textbf{M-AGLO} refers to the layer-by-layer fusing of conditional features \( \textbf{h}_{cond} \) as initially proposed in Section \ref{subsec:Landmark Encoding}. This method also incorporates positional features, represented as \( \hat{\textbf{I}} \), during the MLP phase. It's worth noting that the current methods, including RAD-NeRF \cite{tang2022radnerf} and Geneface++ \cite{ye2023geneface++}, tend to incorporate conditional features into positional features through the use of concatenation. This particular approach closely aligns with the generative latent optimization (GLO) method, as initially introduced in NeRF-w \cite{martinbrualla2020nerfw}. The more recent Zip-NeRF \cite{barron2023zipnerf} has introduced a new feature fusion technique known as affine generative latent optimization (AGLO), which is inspired by the AdaIN \cite{huang2017adain}. M-AGLO specifically adopts AdaIN at each output of the Fully Connected (FC) layer within the MLP. At this phase, the outputs are subjected to an affine transformation.  This specific transformation involves both translation and scaling, which are parameterized by shift \( \alpha \) and scale \( \beta \). These particular parameters are generated from an additional FC branch:
\begin{equation} 
(\alpha, \beta ) = FC(  \textbf{h}_{cond} ),
\end{equation}
\begin{equation} 
\textbf{I} = \hat{\textbf{I}}*(1+\beta) + \alpha, 
\end{equation}
where \( \textbf{I} \) denotes the outputs of M-AGLO. Comparisons between GLO and M-AGLO can be found in Fig. \ref{fig:GLO_and_AGLO}. 

Further improvements can be realized by leveraging the multi-resolution hash grid encoding enabled by progressive optimization.

\begin{table*}
    \setlength{\tabcolsep}{15pt}
    \renewcommand{\arraystretch}{1.1}
    \centering
    \vspace{-0.3cm}
    \begin{tabular}{c  c  c  c  c  c  c}
        \specialrule{0.5pt}{0pt}{0pt}
        \toprule
        Methods & PSNR$\uparrow$ & LPIPS$\downarrow$ & FID$\downarrow$ & LMD$\downarrow$ & Sync$\uparrow$ & Inference FPS$\uparrow$ \\
        \midrule
        Wav2Lip \cite{method2:prajwal2020lip} & N/A  & N/A  & 23.0932   & 2.9506   & \textbf{9.3122}  &  11.95 \\
        RAD-NeRF \cite{tang2022radnerf} & 25.0359   & 0.1326   & 22.0074   & 2.8910  & 6.3348 & 28.37 \\
        ER-NeRF \cite{li2023ernerf} & 26.5732   & 0.1313   & 20.3462   & 2.8131  & 5.5852 & \underline{30.22} \\
        Geneface++\(\dagger\) & \underline{26.6416}   & \underline{0.1069}   & \underline{19.2173}  & \underline{2.3685}  & \underline{7.3568}  & 23.55 \\
        \midrule
        Ours & \textbf{26.9664}   & \textbf{0.0867}   & \textbf{18.5090}  & \textbf{2.2957}    & 7.2962  & \textbf{32.44} \\
        \specialrule{0.5pt}{0pt}{0pt}
        \bottomrule
    \end{tabular}
    \vspace{-0.2cm}
    \caption{Quantitative comparison under the self-driven setting. Our method outperforms other methods in visual generation quality and inference speed. Please refer to Sec. 4.2 for detailed discussions.  We highlight the \textbf{best} and \underline{second best} results.} 
    \label{tab:self-driven}
    \vspace{-0.3cm}
\end{table*}

\noindent \textbf{Conditional Progressive Optimization.} Inspired by the progressive optimizing hash grids in Neuralangelo \cite{li2023neuralangelo}, we propose a conditional progressive optimization scheme. Although the effectiveness of this progressive optimization approach was sufficiently demonstrated by Neuralangelo, given the additional conditions in our task, the coarse-to-fine optimization could not be applied without modifications. The original coarse-to-fine training method results in dynamic conditioning of positional features, which leads to unconvergence during model training. Thus, we introduce conditional progressive optimization based on M-AGLO. Specifically, we replace concatenations with affine transformations applied on the intermediate layers of the MLP. As mentioned earlier, the affine transformation is parameterized by a pair of \( \alpha \) and \( \beta \) generated by a corresponding bypass FC layer. More details can be found in Fig. \ref{fig: Conditional Progressive Optimizationl}.

The advantages of progressive multilayer conditioning can be easily observed in Table \ref{tab: method ablation study}. 

And it is noteworthy that, unlike prior works such as RAD-NeRF \cite{tang2022radnerf} and Geneface++ \cite{ye2023geneface++}, we employ the conditional features as a global conditioning for all positional features. \textit{Such global conditioning eliminates the heavy computations required by audio-spatial decomposition in subsequent processing}, which speeds up the rendering process (as shown in Table \ref{tab:self-driven}). 
\begin{table}
    \setlength{\tabcolsep}{8pt}
    \renewcommand{\arraystretch}{1.1}
    \centering
    \begin{tabular}{ccc}
    \specialrule{0.5pt}{0pt}{0pt}
    \toprule
    Methods   & Cross Gender & Cross Lingual \\
    \midrule
    RAD-NeRF \cite{tang2022radnerf}  &  6.1688   &  3.7548  \\
    ER-NeRF \cite{li2023ernerf}  &   6.1288  &  3.3855   \\
    Geneface++\(\dagger\)  &  \underline{6.4868}   &  \underline{4.0320}   \\

    \midrule
    Ours     & \textbf{6.5965}  & \textbf{4.1113}   \\
    \specialrule{0.5pt}{0pt}{0pt}
    \bottomrule
    \end{tabular}
    \vspace{-0.2cm}
    \caption{Quantitative comparison under the cross-driven setting. Our method outperforms best, and we highlight the \textbf{best} and \underline{second best}.}
    \label{tab:cross-driven}
\end{table}
\subsection{Training Details}
\textbf{Implementation Details.} During the optimization process, the number of levels of the hash grids is set to 16, and the resolution is from \(2^5\) to \(2^{11}\). The maximum number of hash entries for each level is \(2^{16}\). At the beginning of the optimization, we activated the first four layers of the low-resolution hash grid, for every subsequent 5000 iterations, we gradually enabled a new hash grid. We choose AdamW as the optimizer and 240K iterations are performed. The learning rate is reduced by 0.1 and 0.01 times at 150K and 180K respectively. Lips fine-tuning starts at 200K and the learning rate is reset to its initial value.

\noindent \textbf{Maximum Occupancy Grid Pruning.} To enable efficient inference, we adopt the maximum occupancy grid pruning proposed by RAD-NeRF \cite{tang2022radnerf}. For talking heads, occupancy variations under conditioning are usually small, we maintain the occupancy with maximum density values for all conditions during training. 

\noindent \textbf{Loss function.} We use MSE to calculate the color \(\textbf{C}\) loss of pixels as the vanilla NeRF \cite{mildenhall2020nerf}:
\begin{equation} 
\mathcal{L}_{color} = \sum\limits || \textbf{C} - \textbf{C}_{gt}||^2_{2}.
\end{equation} 

In the lips fine-tuning stage, the LPIPS \cite{zhang2018unreasonable} loss is included:
\begin{equation}  
\mathcal{L}_{lip} = \sum\limits || \textbf{C} - \textbf{C}_{gt}||^2_{2} + \lambda LPIPS(\mathcal{P}, \mathcal{P}_{gt}),
\end{equation} 
where \(\mathcal{P}\) is a sampled patch containing the lip and \( \lambda = 0.01\).

\noindent \textbf{Pre-trained Motion Generator} Based on the code of the audio-to-motion and corresponding pre-trained parameters of Geneface \cite{ye2023geneface}, we reproduce the motion generator.

%% file: sec/2_experiments.tex
\section{Experiments}
\label{sec:Experiments}

\begin{table*}
    \setlength{\tabcolsep}{10pt}
    \renewcommand{\arraystretch}{1.1}
    \centering
    \vspace{-0.3cm}
    \begin{tabular}{c  c  c  c  c  c  c}
        \specialrule{0.5pt}{0pt}{0pt}
        \toprule
        Methods & Wav2Lip \cite{method2:prajwal2020lip} & RAD-NeRF \cite{tang2022radnerf} & ER-NeRF \cite{li2023ernerf} & Geneface++\(\dagger\)  & Ours & \\
        \midrule
        Lip-sync Accuracy & \textbf{3.73}  & 3.15  & 3.01   & 3.16   & \underline{3.53}  & \\
        Image Quality & \underline{3.60}   & 3.09  & 3.15   & 3.37  & \textbf{4.12} & \\
        Video Realness & 2.84   & 2.99   & 2.90   & \underline{3.00}  & \textbf{3.56} & \\

        \specialrule{0.5pt}{0pt}{0pt}
        \bottomrule
    \end{tabular}
    \vspace{-0.1cm}
    \caption{\textbf{User Study.} The rating is on a scale of 1-5, the higher the better. We highlight the \textbf{best} and \underline{second best} results.}
    \label{tab:user study}
\end{table*}
\subsection{Experimental Setup}
 


\textbf{Datasets.}
We follow the data preprocessing method of Geneface \cite{ye2023geneface} to process voice and video data. For the person-specific video, to compare with the state-of-the-art method, we still use the dataset in Geneface \cite{ye2023geneface}.

\noindent \textbf{Comparison Baselines.} We compare our method with some classic works: 1) Wav2Lip \cite{method2:prajwal2020lip}: based on the adversarial network architecture, the accuracy of lip-synchronization is improved through pre-trained sync-expert. 2) RAD-NeRF \cite{tang2022radnerf}: inherits the framework of AD-NeRF \cite{guo2021adnerf} and adopts the latest NeRF method instant-NGP \cite{mueller2022instant} to improve the quality of generated results and inference speed. 3) Geneface++ \cite{ye2023geneface++}: mapping audio to facial 3D landmarks with a motion generator, and then input the statically filtered facial 3D landmarks as a condition into NeRF. Because the motion generator has good generalizability, it improves audio-lip synchronization. Since the Geneface++ code has not been open-sourced yet, we temporarily use the audio2motion code of Geneface and instant-NGP based NeRF module of Geneface++ to ensure the fairness of comparison to the greatest extent. We denote the replicated one as Geneface++\(\dagger\). \textit{We look forward to conducting complete comparative experiments with Geneface++ once its code is released in the near future.} 4) ER-NeRF \cite{li2023ernerf}: Tri-Plane hash representation is introduced to improve the accuracy of head rendering, and the attention mechanism is used to decouple the alignment relationship between audio and space. \textit{To ensure the fairness of comparison, we apply the LPIPS loss only in the lip area, like others, in the lips fine-tuning stage of ER-NeRF.} Due the space limitation, we have put more descriptions in the supplementary material.




\begin{figure}
    \centering
    \includegraphics[width=1.0\linewidth]{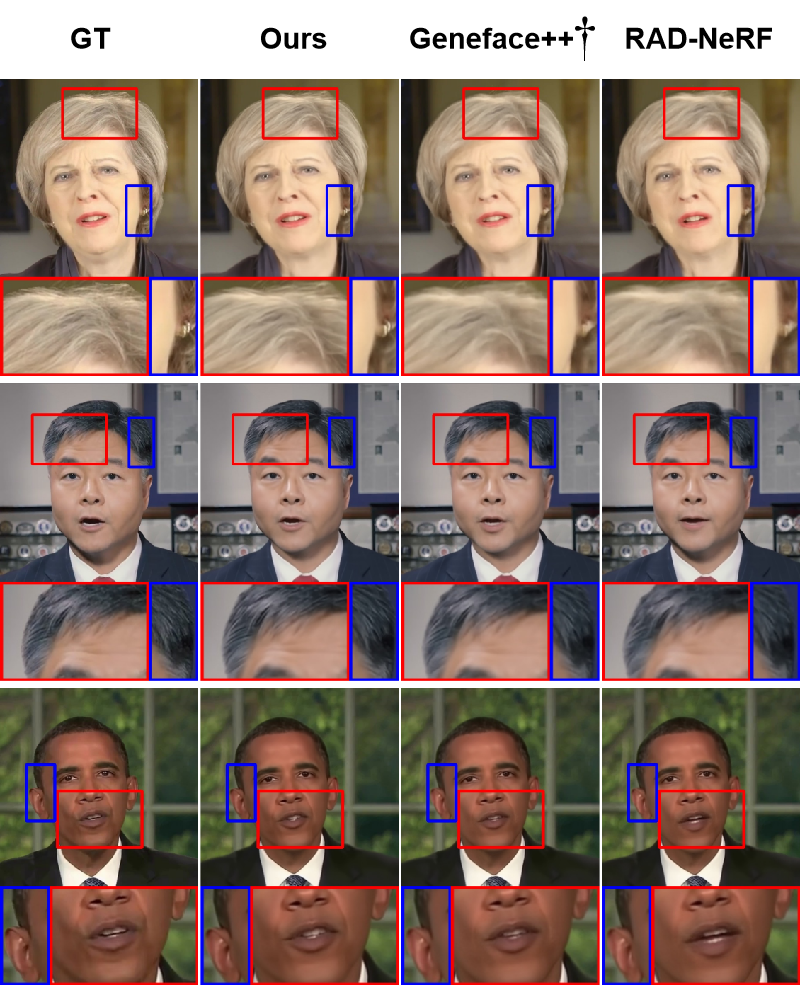}
    \vspace{-0.5cm}
    \caption{\textbf{Self-driven Quality Comparison.} Under the self-driven settings, we visually compare the visual quality of the results generated by different methods. It can be seen that our method can generate more details and more fidelity results. Please zoom in for better visualization.}
    \label{fig:self driven image quality}
\end{figure}
\subsection{ Quantitative Evaluation}

\textbf{Metrics.}
We use PSNR, LPIPS \cite{zhang2018perceptual} and FID \cite{Seitzer2020FID} to measure image quality. We also adopt LMD \cite{chen2018lmd} and Sync (SyncNet confidence score) \cite{Chung16Syncnet}  to measure lip synchronization.

\noindent \textbf{Comparison Setting.}
Like RAD-NeRF, we perform comparisons under two settings. 1) \textit{self-driven}: dividing the dataset of the same individual into a training set and a test set. In this case, we have the corresponding ground truth, so we can use PSNR, LPIPS and FID to evaluate the image quality. Both LMD and Sync are used to evaluate the accuracy of audio-lip synchronization. 2) \textit{cross-driven}: using out-of-domain audio to drive the generation. The driving dataset in Geneface \cite{ye2023geneface} is used here. In this case, we can use Sync to evaluate the accuracy of lip synchronization.

\noindent \textbf{Evaluation Results.}
Under the self-driven setting, our method performs best in all metrics and has significant improvements, except the Sync (as shown in Table \ref{tab:self-driven}). However, our method is still comparable to Geneface++ \cite{ye2023geneface++} in Sync. Since Wav2Lip \cite{method2:prajwal2020lip} involves the expert model of Sync during training, the score ahead a large margin to the second place. On the other hand, since the results generated by Wav2Lip \cite{method2:prajwal2020lip} need to be warped back to the ground truth, it is unreasonable to measure them with PSNR and LPIPS, so we set them to N/A. As for inference speed, due to the absence of source codes for some methods and constraints on our resources, we conduct speed tests for RAD-NeRF and our proposed method. Subsequently, we perform a rational proportional conversion of the results and the ones mentioned in Geneface++ \cite{ye2023geneface++}. Due to the space limitation, we have put more detail and add more discussions in the supplementary material.

The cross-driven evaluation results are shown in Table \ref{tab:cross-driven}. Under the settings of cross-gender and cross-lingual, our method performs best in terms of the Sync, followed by Geneface++ \cite{ye2023geneface++}. The experiment results show that our proposed method has better generalization, which corroborates the results of the previous analysis Section \ref{subsec:Landmark Encoding}. 


\begin{figure*}
    \centering
    \vspace{-0.3cm}
    \includegraphics[width=1\linewidth]{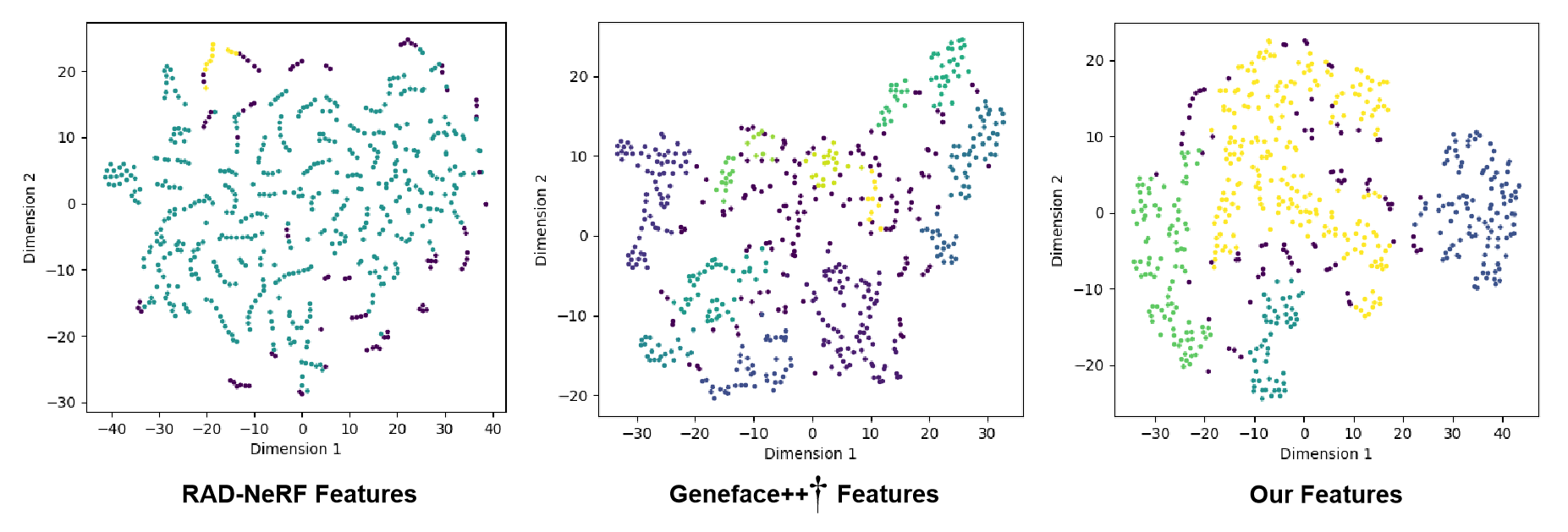}
    \vspace{-0.7cm}
    \caption{\textbf{Conditional Feature Visualization.} Compared with others, our landmark encoder can obtain features with better separability and aggregation.}
    \label{fig: Conditional Feature Visualization}
    \vspace{-0.2cm}
\end{figure*}


\subsection{Qualitative Evaluation}
\noindent \textbf{Evaluation Results.}
For self-driven, Fig. \ref{fig:self driven image quality} compares the generated image quality of three different portrait videos driven by different methods. By zooming in on the details, it can be seen that our proposed method can restore more details and bring a better visual experience. For cross-driven, in Fig. \ref{fig:cross-driven image quality}, We also compare the performance under the cross-driven settings. Our proposed method can generate more accurate synced lip and realistic results than Geneface++ \cite{ye2023geneface++} and RAD-NeRF \cite{tang2022radnerf} which is consistent with the cross-driven quantitative evaluation results.
 
\noindent \textbf{User Study.} We introduce user studies to evaluate the generation quality of various methods further. We randomly invited 25 volunteers to participate in the study. Volunteers evaluated 15 videos generated by different methods under the cross-driven setting. Evaluation dimensions include: 1) audio-lip synchronization (\textbf{Lip-sync Accuracy}), 2) image generation quality (\textbf{Image Quality}), and 3) overall video realness (\textbf{Video Realness}). Each dimension is scored on a scale of 1 to 5 (higher scores mean better performance). We use the mean opinion score (MOS) protocol for evaluation. 


As can be seen in Table \ref{tab:user study}, our method received the highest scores in image quality and video realness. As mentioned earlier, since the results of Wav2Lip \cite{method2:prajwal2020lip} are warped back to the original image, the volunteers believed that the image quality of Wav2Lip takes the second place. However, since this warped back will cause obvious synthesized artifacts, Wav2Lip has the lowest score in video realism. Geneface++ \cite{ye2023geneface++} is the second only to ours in video realness. Finally, Wav2Lip is the best at Lip-sync Accuracy, followed by ours. This suggests that the synthesized lips by Wav2Lip \cite{method2:prajwal2020lip} are also the most subjectively in line with the expectations of people. At the same time, under the user study setting, our score is also closer to Wav2Lip \cite{method2:prajwal2020lip}, which is more in line with the data distribution expectation. It also surpasses Geneface++ \cite{ye2023geneface++}, further demonstrating that our method has better generalization ability.

\subsection{Ablation Study}
\noindent \textbf{Conditional Feature Visualization.} Figure \ref{fig: Conditional Feature Visualization} presents the visualizing of the conditional features between RAD-NeRF \cite{tang2022radnerf}, Geneface++ \cite{ye2023geneface++}, and our proposed method by T-SNE \cite{van2008visualizing}. It is evident that RAD-NeRF's feature separability, which relies on the audio extractor, is relatively poor. While both Geneface++ and our approach extract features based on 3D landmarks, our method, which losslessly encodes conditional features based on the spatial positions of 3D landmarks, exhibits superior separability and aggregation. This aligns with our expectations (as discussed in Section \ref{subsec:Landmark Encoding}), and is consistent with our prior quantitative evaluation results, further demonstrating the effectiveness of our approach.

\noindent \textbf{Method Ablation Study.}
We analyze the performance improvement of the various modules introduced in Table \ref{tab: method ablation study}. Employing RAD-NeRF as the initial baseline, we incrementally incorporate Landmark Encoding (Section \ref{subsec:Landmark Encoding}), M-AGLO (Section \ref{subsec:Progressive Multi-Layer Conditioning}), and Progressive Optimization (Section \ref{subsec:Progressive Multi-Layer Conditioning}). With the incorporation of Landmark Encoding, a significant improvement is observed across all metrics when compared to the baseline. Subsequently, with the adoption of M-AGLO, there is a notable further enhancement in LPIPS \cite{zhang2018perceptual}. Lastly, with the integration of Progressive Optimization, we observe additional improvement in PSNR, while other performance indicators do not exhibit significant changes. Further details can be found in Table \ref{tab: method ablation study}.

\begin{table}
    \setlength{\tabcolsep}{2pt}
    \renewcommand{\arraystretch}{1.1}
    \centering
    \begin{tabular}{c  c  c  c  c  c}
        \specialrule{0.5pt}{0pt}{0pt}
        \toprule
         & PSNR$\uparrow$ & LPIPS$\downarrow$  & LMD$\downarrow$ & Sync$\uparrow$ & \\
        \midrule
          RAD-NeRF \cite{tang2022radnerf}                  & 25.0359   & 0.1326    & 2.8910  & 6.3348   & \\
        + Landmark encoding         & 26.5861   & 0.0939    & 2.3311  & 7.1664   & \\
        + M-AGLO                    & \underline{26.7414}   & \textbf{0.0865}    & \textbf{2.2949}  & \underline{7.2800}   & \\
        + Progressive optimization  & \textbf{26.9664}   & \underline{0.0867}    & \underline{2.2957}  & \textbf{7.2962}  & \\
        \bottomrule
    \end{tabular}
    \vspace{-0.2cm}
    \caption{\textbf{Ablation Study on Modules.} We highlight the \textbf{best} and \underline{second best} results}
    \label{tab: method ablation study}
\end{table}

\section{Ethical Consideration}
Our method can synthesize high-fidelity, real-time talking head video, trained on a video of several minutes in length. We aspire to apply this approach to endeavors that are advantageous for society. As part of our commitment, we are eager to offer support and assistance to the deepfake detection community. We firmly believe that our method if utilized judiciously can contribute to the wholesome advancement of digital human technology.

%% file: sec/3_finalcopy.tex
\section{Conclusion}
In this paper, we present R2-Talker, an efficient and effective framework for realistic real-time talking head synthesis. We propose a structure-aware 3D facial landmark encoder for extracting more precise and generalized conditional features. We also propose a progressive conditioning approach for compactly fusing conditional features and positional features. Extensive quantitative and qualitative experiments show that our method outperforms the existing stat-of-the-art methods with more realistic, high-fidelity, audio-lip synchronization and real-time inference.

%% file: sec/X_suppl.tex
\clearpage
\setcounter{page}{1}
\maketitlesupplementary
\section*{A. Overview}
\label{sec:Overview}
In the supplemental document, we introduce the details of the baseline settings, the profile of inference speed, and the ablation study of the level dim of the hash grid in R2-Talker. For \textit{details on comparison baselines}, please refer to baseline settings in Section B. And for the\textit{ details of inference speed} in quantitative evaluation, please refer to the profile of inference speed in Section C. Finally, our project page can be visited at \href{https://kylinyee.github.io/R2-Talker/}{R2-Talker} which is under construction and can be accessed around Dec 25, 2023, Pacific Time.

\section*{B. Baseline Settings}
\label{sec: Baseline Settings}
In our study, we compare R2-Talker with four classic approaches: Wav2Lip, RAD-NeRF, ER-NeRF, and Geneface++.

\noindent \textbf{Wav2Lip \cite{method2:prajwal2020lip}.} We utilize the code repository available at \href{https://github.com/Rudrabha/Wav2Lip.git}{https://github.com/Rudrabha/Wav2Lip.git}  and employ the pre-trained Wav2Lip+GAN model with default settings.

\noindent \textbf{RAD-NeRF \cite{tang2022radnerf}.} We use the open-source code from \href{https://github.com/ashawkey/RAD-NeRF.git}{https://github.com/ashawkey/RAD-NeRF.git}. After reviewing the source code and settings, we directly employ the default settings for model training.

\noindent \textbf{Geneface++ \cite{ye2023geneface++}.} As Geneface++ is not open-source, we refer to the code repository of Geneface at \href{https://github.com/yerfor/GeneFace.git}{https://github.com/yerfor/GeneFace.git}. After reading and testing the relevant codes, we implement an approximation of Geneface++, named Geneface++\(\dagger\), which is based on the motion generator of Geneface \cite{ye2023geneface} and using instant-NGP \cite{mueller2022instant} as the NeRF render module. Since the computational increase of the motion generator of Geneface++ over that of Geneface is negligible in terms of speed, we refer to the inference speed of Geneface++ as equal to Geneface++\(\dagger\), namely Geneface++\( \dagger \) = Geneface+instant-NGP. Furthermore, according to Geneface++, the enhancements in the motion generator, specifically Pitch-Aware and landmark LLE optimizations, contribute to improving generation results. Therefore, the generated quality and lip-sync accuracy of Geneface++\(\dagger\) is expected to be slightly inferior to Geneface++. We look forward to conducting complete comparative experiments with Geneface++ once its code is released in the near future.

\begin{table*}
    \setlength{\tabcolsep}{4pt}
    \renewcommand{\arraystretch}{1.1}
    \vspace{-0.5cm}
    \centering
    \begin{tabular}{ccccccc|c}
    \specialrule{0.5pt}{0pt}{0pt}
    \toprule
       ~ & Wav2vec2.0 & HuBERT & Audio2motion & Render module & Our render module & total time@450x450 & FPS \\
       \midrule
       RAD-NeRF                  &  1.64 &   -   &   -  & 21.95 &   -   & 23.59 & 42.39\\
       Geneface++\(\dagger\)	  &   -   &  4.62 & 0.06 & 21.95 &   -   & 26.63 & 37.55\\
       Ours	                  &   -   &  4.62 & 0.06 &   -   & 14.66 & 19.34 & 51.71\\
       \specialrule{0.5pt}{0pt}{0pt}
       \bottomrule
    \end{tabular}
    \caption{We test the inference speed of the individual modules of the NeRF-based approaches. The modules include 1) audio extraction and processing modules: Wav2vec2.0, HuBERT, and Audio2motion. 2) generation modules: Render module and Our proposed render module. According to the speed of different modules, the overall time consumption of different methods is calculated. It can be seen that our method is the fastest. Like Geneface++\( \dagger \), HuBERT+Audio2motion has been introduced to our method as an audio extraction and processing module. However, with the efficient render module, the overall time of our method is only 19.34ms.}
    \label{tab: profile of inference speed}
\end{table*}

\begin{table*}
    \centering
    \begin{tabular}{lll}
    \specialrule{0.5pt}{0pt}{0pt}
    \toprule
    Modules & Hyper-parameter & Value \\
    \midrule
    \multirow{6}{*}{Hash Grids based Positional Encoder}  & Input dim & 3\\
    ~     & Number of levels & 16\\
    ~     & Level dim	 & 4\\
    ~     & Base resolution & 32\\
    ~     & Log2 hashmap size & 16\\
    ~     & Desired resolution & 2048\\
    \midrule
    \multirow{4}{*}{Sigma Net} & Input dim & 4*16\\
    ~     & Layers & 3\\
    ~     & Hidden dim	 & 128\\
    ~     & Output dim & 128+1 \\
    \midrule
    \multirow{6}{*}{Hash Grids based Landmark Encoder}  & Input dim & 3\\
    ~     & Number of levels & 16\\
    ~     & Level dim	 & 2\\
    ~     & Base resolution & 32\\
    ~     & Log2 hashmap size & 16\\
    ~     & Desired resolution & 2048\\
   \midrule
   \multirow{4}{*}{FCs in progressive conditioning} & Input dim & 5*landmark num*Level dim*Number of levels\\
    ~     & Layers & 3\\
    ~     & Hidden dim	 & 128\\
    ~     & Output dim & 2*hidden dim of target intermediates\\
    \specialrule{0.5pt}{0pt}{0pt}
    \bottomrule
    \end{tabular}
    \caption{The hyperparameter configurations of the two modules which are proposed in our method are listed, for the reader to further understand the details of our experiments. Including Hash grids based Positional Encoder, Sigma Net,  Hash grids based Landmark Encoder, and FCs in progressive conditioning. For the other modules, we keep the same settings as the previous works.  The additional number of landmarks, dimension of landmarks, and hidden dimension of target intermediates are set as \( 68, 3, 64\) respectively.}
    \label{tab: Hyper params}
\end{table*}

\noindent \textbf{ER-NeRF \cite{li2023ernerf}.} In the main paper, to ensure the fairness of comparison, we apply the LPIPS loss only in the lip area in the lips fine-tuning stage of ER-NeRF. Here we present more details about the implementations. We use the open-source code from \href{https://github.com/Fictionarry/ER-NeRF.git}{https://github.com/Fictionarry/ER-NeRF.git} and thoroughly review the code and settings. We note that the training of ER-NeRF is also divided into two phases, like previous works. The two phases include head training and torso training. The head training involves two steps: the first step optimizes the model supervised on the overall face with L2 loss, and the second step, lip-finetuning, optimizes the local lip area. In the second step, global face and local lip optimizations are alternately optimized based on iteration steps, employing both L2 loss and Lpips loss. This loss setting in the second step differs from the other methods. The following Table \ref{tab: ER-NeRF lip-fine-tuning settings} provides a more clear comparison. Concerning the significant improvements of Lpips loss on generated results, to ensure objectivity and fairness in comparison, we adjusted the lip-finetuning training settings of ER-NeRF to align with the other methods mentioned, i.e., setting the training loss for global face to only L2 loss, while maintaining L2 loss and Lpips loss for local lip training. Other settings are kept as default.
\begin{table}
    \setlength{\tabcolsep}{5pt}
    \renewcommand{\arraystretch}{1.1}
    \centering
    \begin{tabular}{lll}
        \specialrule{0.5pt}{0pt}{0pt}
        \toprule
        Method & global face & local lip\\
        \specialrule{0.5pt}{0pt}{0pt}
        \midrule
        RAD-NeRF/Geneface++\( \dagger \) & L2   & L2+Lpips \\
        ER-NeRF & L2+Lpips  & L2+Lpips \\
        \specialrule{0.5pt}{0pt}{0pt}
        \bottomrule
    \end{tabular}
    \caption{The table compares the loss settings of ER-NeRF and other methods in the global face and local lip stage. In the local lip stage, ER-NeRF also introduces Lpips loss on the basis of L2 loss. Lpips loss has a great impact on the final result. So for a more objective comparison, in the only lip stage, we turned off the Lpips loss of ER-NeRF.}
    \label{tab: ER-NeRF lip-fine-tuning settings}
\end{table}

\noindent \textbf{Our approach.} We adopt the motion generator from Geneface \cite{ye2023geneface}, with the NeRF module using our designed framework. The loss function settings for head training and torso training phases are kept consistent with RAD-NeRF and Geneface++. The Table \ref{tab: Hyper params} below further details the hyper-parameters of some important modules in our approach.

\section*{C. Profile of Inference Speed}
\label{sec: Profile of Inference Speed}
Due to graphics card scheduling and time constraints, we did not report the results of inference testing on RTX3090 in the main paper. Instead, we test the inference time of RAD-NeRF and our method on A100 series GPU and then refer to the test data in the paper of Geneface++ \cite{ye2023geneface++} and obtain a reference result through proportional conversion. Here we show the test results and analysis details on RTX3090, with results in Table \ref{tab: profile of inference speed} (measurements are in milliseconds except for FPS). The NeRF-based audio-driven talking head synthesis method usually consists of two parts, audio extraction and render module. Based on this structure, we test the inference speed of the two parts separately. And we select RAD-NeRF, Geneface++\( \dagger \), and our method for comparison. Because these three methods have similar modules, our comparative analysis is performed in a more fine-grained manner, i.e., the module-level profile of inference speed.

Firstly, the time consumption of RAD-NeRF includes two parts: audio feature extraction using wav2vec2.0 and rendering with the render module, totaling 23.59ms, achieving 42.39 FPS. The rendering module in RAD-NeRF is based on the audio-spatial decomposition method implemented with instant-NGP. The first part of Geneface++\( \dagger \) includes audio extraction using Hubert and the audio2motion module to convert audio features into 3D facial landmarks, followed by the render module from RAD-NeRF, totaling 26.63ms. Our method adopts the approach of Geneface++\( \dagger \) for the audio to landmarks and the more efficient and compact render module for the second part. Specifically, our render module is based on hash grid landmark encoding and progressive conditioning. Thus, even with a longer duration in the first part, the overall time consumption is only 19.34ms, reaching the highest frame rate of 51.71 FPS.
\begin{table}
    \centering
    \setlength{\tabcolsep}{3pt}
    \renewcommand{\arraystretch}{1.1}
    \begin{tabular}{c|lllll}
    \specialrule{0.5pt}{0pt}{0pt}
    \toprule
       Level dim  & PSNR$\uparrow$ & LMD$\downarrow$ & LPIPS$\downarrow$ & FID$\downarrow$ & Sync$\uparrow$ \\
       \midrule
       2  & \underline{28.7105} & \underline{2.6213} & 0.1000 & 17.2397 & \underline{8.239} \\
       4  & 28.6665 & \textbf{2.5505} & \underline{0.0918} & \underline{17.1687} & 8.131 \\
       8  & \textbf{28.8402} & 2.6424 & \textbf{0.0901} & \textbf{16.7103 }& \textbf{8.317} \\
    \specialrule{0.5pt}{0pt}{0pt}
    \bottomrule
    \end{tabular}
    \caption{As seen in the table, when the level dim changes from 2 to 8, the overall level dim is positively correlated with the increase in each indicator. These gains are relatively weak, thus we do not believe that these gains will contribute to the core improvements of our approach. Therefore, it can be shown that setting the hyper-parameter of level dim to 4 is acceptable. And we highlight the \textbf{best} and \underline{second best} results.}
    \label{tab: Ablation Study of Level Dims}
\end{table}

\section*{D. Ablation Study of Level Dim}
\label{sec: Ablation Study of Level Dim}
As shown in Table \ref{tab: Hyper params}, the hash grids based on the positional encoder and sigma net in our method are almost directly applied with the settings of previous methods such as RAD-NeRF and Geneface++. The level dim in the hash grids based positional encoder is set to 4 in our method, whereas the previous method is usually set to 2. This setting is consistent with the Neuralangelo \cite{li2023neuralangelo}. Further, we conduct ablation experiments to validate the influence of this parameter without significant improvements observed. So, our setting of the level dim is reasonable. More details can be found in Table \ref{tab: Ablation Study of Level Dims}.